# Physics-Guided Deep Unfolding for Blind Cross-Sensor Spectral Super-Resolution via Learning the Spectral Transformation Function

Zhaolin Li[a,b]; Jinsong Chen[a,c*]; Shanxin Guo[a,c*]; Tuo Zhang[a,b]; Xinglong Zhang[a,b]; Pan Chen[a,c]

*Abstract*— **Hyperspectral imaging provides rich spectral information for quantitative remote sensing, yet hyperspectral sensors remain costly and thus unavailable in many UAV deployments. Spectral super-resolution (SSR) seeks to reconstruct hyperspectral images (HSIs) from multispectral images (MSIs). Most existing SSR methods assume a fixed and known spectral response function (SRF) and are therefore limited to single-sensor settings. In practical cross-sensor scenarios, the spectral degradation from HSI to MSI is unknown and varies with sensor characteristics and scene content, which renders HSI reconstruction ill-posed. This paper proposes a physics-guided deep unfolding network, termed PGU-Net, to address blind cross-sensor SSR by jointly estimating the HSI and a learnable spectral transformation function (STF). PGU-Net unrolls an alternating optimization procedure into an end-to-end trainable architecture with $K$ stages, where each stage sequentially updates the HSI and the STF. Both modules combine learnable proximal networks with differentiable closed-form solvers, enabling physical interpretability while retaining strong representation capacity. Experiments on benchmark datasets (CAVE and NTIRE 2022) with multiple SRFs demonstrate accurate recovery of the STF (degradation operator) and improved reconstruction performance over state-of-the-art SSR methods. Furthermore, evaluations on a real UAV cross-sensor dataset (Headwall Nano HSI and DJI P4 Multispectral MSI) verify the effectiveness and robustness of PGU-Net under truly blind conditions, and suggest that the estimated STF may exhibit land-cover-related differences.**

## I. Introduction

With the rapid growth of UAV remote sensing, UAV-based spectral imaging systems have matured significantly. Compared with multispectral images (MSI), hyperspectral images (HSIs) provide richer spectral details that enable applications ranging from precision agriculture to accurate environmental monitoring [1],[2],[3]. However, the high cost of hyperspectral sensors and the large data volume of HSIs limit their widespread deployment. Conversely, MSI sensors are more affordable and ubiquitous. However, due to their limited spectral resolution, MSI sensors often lack the fidelity required for complex analysis, limiting their utility in quantitative retrieval tasks [4]. Spectral super-resolution (SSR) aims to reconstruct HSIs from MSI observations, improving spectral fidelity for downstream quantitative tasks [5].

Based on different mapping approaches, SSR can be broadly categorized into three categories: 1) physics-based matrix decomposition, 2) data-driven nonlinear end to end modeling, and 3) hybrid unrolling model that combines both physical and data-driven approaches.

Physics-based matrix decomposition was developed earliest and primarily involves describing how narrow-band signals are combined into broad-band signals using point spread functions (PSFs) and spectral response functions (SRFs). PSF affects spatial mixing, which indirectly impacts the effective spectral observation at coarser scales [6]. The transformation from broad-band to narrow-band signals is then accomplished through matrix decomposition or sparse representation techniques [5]. Early research attempted to optimize the linear decomposition models to achieve this goal [7]. Examples of such algorithms include Principal Component Analysis (PCA) [8], Karhunen–Loève transformation [9] and Pseudo-Inverse (PI) [10]. However, the relationship between broad-band and narrow-band spectra cannot always be accurately captured by linear models. To address this limitation, nonlinear models such as sparse representation [11], [12], Gaussian Processes [13], and manifold learning [14] were introduced. But, the dependence of these physics-based methods on predefined unmixing processes inherently compromises their robustness to noise and their flexibility in handling large, spectrally variable datasets, typically leading to marked reductions in accuracy.

With the development of deep neural networks, data-driven spectral super-resolution techniques have rapidly evolved, enabling end-to-end mapping to be directly established between multispectral and hyperspectral data. These models utilize deep neural layers to dynamically fit the nonlinear mapping functions to enhance the accuracy of unmixing in challenging scenarios. One idea is to consider defining the spectral unmixing problem as deconvolution processing. Representative approaches use a Convolutional Neural Network (CNN)-based deconvolution structure to unmix the broadband to the narrow

The authors are with [a] Center for Geo-Spatial Information, Shenzhen Institutes of Advanced Technology, Chinese Academy of Sciences, Shenzhen, 518055, PR China; [b] University of Chinese Academy of Sciences, Beijing 100049, China; [c] Shenzhen Engineering Laboratory of Ocean Environmental Big Data Analysis and Application, Shenzhen, 518055, PR China. (Corresponding author: Jinsong Chen. Email: js.chen@siat.ac.cn; Shanxin Guo. Email: sx.guo@siat.ac.cn)

band [15], while other models define the unmixing process as a progressive spectral detail enhancement process using 2D/3D convolution [16], [17] and joining the attention model [18], [19]. In addition, degradation models and generative models in the computer vision community field have also been introduced to solve the spectral super-resolution problem. These models include the Generative Adversarial Network (GAN)-based model [20], the transformer-based model [21], and the diffusion model (DPM) [22], [23]. However, the reconstruction of spectral features by end-to-end deep neural networks requires a sufficient number of training samples. When the distribution of samples is limited, the model is likely to perform overfitting and may generalize poorly across different SSR scenarios (e.g., cross-date or cross-scene transfer). Moreover, their black-box nature complicates error source evaluation and uncertainty assessment, hindering error traceability across diverse data.

To address the poor interpretability and generalization ability of data-driven models, research has demonstrated the importance of incorporating spectral physical constraints into deep networks [24]. One of the ideas is using known SRFs as guidance to simulate the spectral degradation process to add consistency constraints for spectral super-resolution network [25], [26]. This specific constraint is that the multispectral images that have been degraded from the recovered hyperspectral images should be consistent with the original multispectral input images. The key to build the spectral degradation process is to synthesize MSI from HSI by involving a given sensor's SRFs in consideration. Arad et al. [27] demonstrate that incorporating this simulated spectral degradation process could improve spectral prediction accuracy by over 33%. Fu et al. [28] propose a SRF selection layer, which can automatically determine the optimal SRF from the network weights. Martínez [29] demonstrated that SSR models can be effectively pretrained using RGB images when the degradation process and SRFs are known. Li et al. [18] further extended this idea to satellite imagery (GF1 and GF6) with a progressive spatial–spectral joint network (PSJN) [30]. These studies highlight the crucial role of physical spectral degradation process in guiding network learning processes. In this hybrid structure, physical spectral degradation process encourages physically consistent hyperspectral predictions with the known SRFs, and the neural networks were used as regularization constraint terms, or residual terms in the spectral degradation model to maintain the nonlinear modeling capabilities.

Such degradation models must be guided by a known degradation matrix. Therefore, the current hybrid unrolling model are most built for the single sensor SSR problem where the MSIs are simulated by the given HSIs with the given SRF. In these controlled simulation studies, the SRF is assumed to be known and fixed. However, in real-world applications, the SRF is often unknown since manufacturers do not disclose spectral sensitivities. In addition, it can also be influenced by factors such as sensor aging and radiation conditions[31], [32]. This problem is further compounded in cross-sensor SSR problem. In these scenarios, the transformation spectral process between HSI and MSI sensors involves the spectral mixing at the different observation level and the SRFs of sensors. Therefore, no explicit degradation matrix is available to guide the process, and the overall cross-sensor spectral transformation becomes unknown[33].

To extend physics-based degradation models from single-sensor to cross-sensor contexts, in this study, we investigate whether the effective cross-sensor spectral transformation can be learned directly from paired MSI–HSI data in cross-sensor settings. This approach offers two key advantages: (1) the deep learning model can be guided by physical principles of spectral mixing and spectral response processes, leading to more accurate predictions; and (2) the spectral mixing function can be explicitly modeled through end-to-end training, providing new insights into how light from different land cover types mixes across varying observational scales.

Motivated by this, we propose a novel framework based on the physics-guided deep unfolding paradigm, which we refer to as the Physics-Guided Unfolding Network (PGU-Net). This approach preserves the interpretable structure of traditional optimization algorithms while employing deep networks to learn complex cross-sensor spectral mixing processes. PGU-Net formulates the blind SSR problem as the joint estimation of the HSI and the sensor transformation. We refer to this effective cross-sensor transformation as the Spectral Transformation Function (STF) between two sensors. The underlying alternating optimization process is unfolding into a multi-stage network, enabling end-to-end training to address the challenges of SSR in cross-sensor scenarios. The contributions of this work are as follows:

1. We formulate blind cross-sensor SSR as a joint estimation problem of the HSI X and an explicit spectral transformation matrix R (STF) that models the unknown cross-sensor spectral degradation.

2. We develop PGU-Net, a physics-guided deep unfolding framework that unrolls an alternating optimization algorithm into a multi-stage network, integrating learnable proximal operators with differentiable closed-form solvers for both X and R.

3. Extensive experiments on simulated and real cross-sensor data validate that PGU-Net improves reconstruction accuracy while yielding physically meaningful STF estimates; additional analyses indicate that the estimated STF exhibits consistent variations correlated with land-cover categories.

The remainder of this paper is organized as follows. Section II presents the basic ides of our work and proposed model in detail. Section III describes the experimental data and results. Section IV analyzes land-cover dependency and presents ablation studies. Section V concludes the paper.

## II. Proposed Method

Blind spectral super-resolution (SSR) in cross-sensor settings is inherently ill-posed because the spectral degradation from HSI to MSI is unknown. This section first analyzes the cross-sensor spectral transformation mechanism and defines the Spectral Transformation Function (STF). We then present the Physics-Guided Unfolding Network (PGU-Net), which jointly estimates the target HSI and the STF in an interpretable, end-to-end trainable framework.

### *A. Spectral Mixing Process and Spectral Transformation Function (STF)*

As illustrated in **Fig. 1**(a), the imaging chain comprises two distinct stages: the Spectral Mixing Process (SMP) and the Spectral Response Process (SRP). The SMP mixes spectral energy from various endmembers and is governed by sensor characteristics (e.g., Point Spread Function, PSF) and observation conditions (e.g., land cover composition and BRDF) [34]. Subsequently, the SRP converts this mixed light into electronic signals. Consequently, the observed spectrum is a joint function of the Spectral Mixing Function (SMF) and the Spectral Response Function (SRF), modeled as:

$$HSI = SRF_{hsi} \cdot [SMF_{hsi}(S_1, S_2, \dots, S_h)] \quad (1)$$

$$MSI = SRF_{msi} \cdot [SMF_{msi}(S_1, S_2, \dots, S_h)] \quad (2)$$

Where, $S = \{S_1, S_2, \dots, S_h\}$ denotes the set of endmember spectra, and $h$ is the number of endmembers. $SRF$ and $SMF$ denote the spectral response and mixing functions for the HSI and MSI sensors, respectively.

Assuming both sensors observe the same field of view (i.e., share the same endmembers), MSI can be expressed as a transformation of HSI as in Eq. (3):

$$MSI = SRF_{msi} \cdot SMF_{msi}\left[SMF_{hsi}^{+}\left(SRF_{hsi}^{+}(HSI)\right)\right] \quad (3)$$

Where $(\cdot)^{+}$ denotes a generalized inverse operator used to represent the inverse SRP/SMP in an abstract form.

For cross-sensor SSR, we formally define this mapping as the Spectral Transformation Function (STF). As described in Eq.4, the STF is a composite matrix determined by four components: 1) the forward $SRF_{msi}$; 2) the forward $SMF_{msi}$ ; 3) the inverse $SMF_{hsi}^{+}$ ,and 4) the inverse $SRF_{hsi}^{+}$.

$$STF = SRF_{msi} SMF_{msi} SMF_{hsi}^{+} SRF_{hsi}^{+} \quad (4)$$

Crucially, while $SRF_{msi}$ and $SRF_{hsi}^{+}$ are sensor-dependent and fixed, $SMF_{msi}$ and $SMF_{hs}^{+}$ are scene-dependent, varying significantly with land cover types and observation geometry. Since these scene-dependent terms are unknown, the problem is ill-posed and cannot be solved by simple linear systems. To address this, we propose a learnable module, $R$, within PGU-Net to approximate this STF and guide the spectral degradation process during end-to-end training:

$$MSI = \hat{R}(HSI), \hat{R} \approx STF \quad (5)$$

This formulation allows PGU-Net to explicitly model the spectral transformation from HSI to MSI. Unlike traditional methods that rely on static SRFs, the proposed learnable STF may capture land-cover-related differences in effective mixing, offering a potential tool for further analysis.

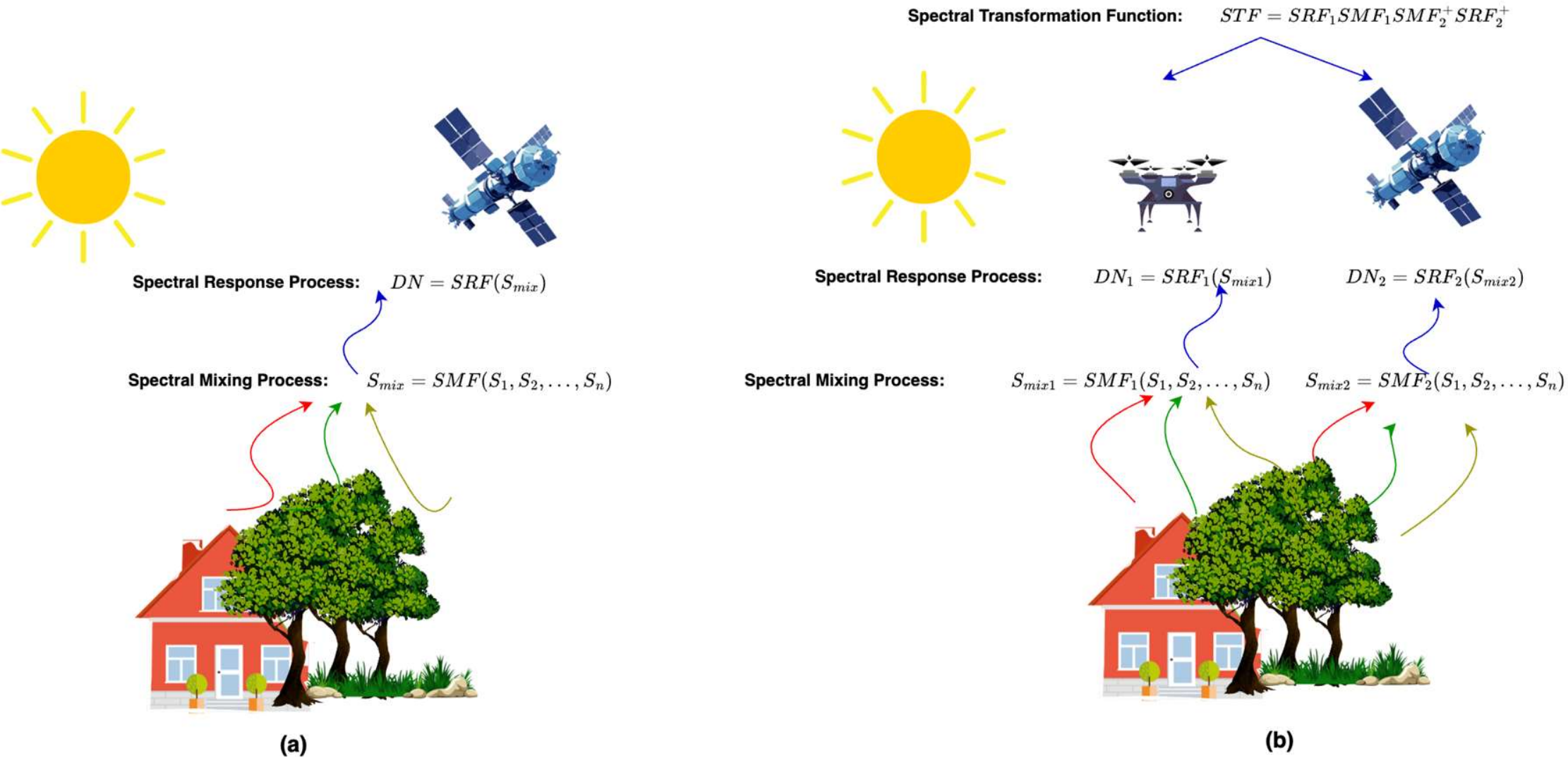

**Fig. 1.** Spectral mixing process on (a) single-sensor scenario and (b) cross-sensors scenario.

### *B. Physical Model and Problem Formulation*

We model the relationship between a hyperspectral image (HSI), denoted as $X \in \mathbb{R}^{W \times H \times C}$ and its corresponding multispectral image (MSI), denoted as $Y \in \mathbb{R}^{W \times H \times c}$, where $W$ and $H$ are the spatial dimensions, $C$ is the number of hyperspectral bands, and $c$ is the number of multispectral bands ($c < C$). The degradation process is modeled as:

$$Y = R(X) \quad (6)$$

Here, $R$ represents the degradation matrix. In controlled single-sensor scenarios, $R$ is typically a known, fixed SRF. However, in blind cross-sensor scenarios, $R$ represents the unknown, scene-dependent STF. Consequently, recovering $X$ becomes a joint optimization problem:

$$\left(\hat{X}, \hat{R}\right) = \underset{X,R}{\operatorname{argmin}} \|Y - RX\|^2 + \tau\varphi(X) + \varepsilon\phi(R) \quad (7)$$

Where $\|Y - RX\|^2$ is the data fidelity term, $\varphi(X)$ and $\phi(R)$ represent the regularization priors for the HSI and STF, respectively; and $\tau$ and $\varepsilon$ are balance parameters. To solve this intractable joint problem, we employ an Alternating Optimization (AO) strategy, decoupling it into two sub-problems (as shown in Fig.2):

1. HSI reconstruction (**SR Module**): Estimate $X$ with fixed $R$.
2. Degradation reconstruction (**STF Module)**: Estimate $R$ with fixed $X$.

PGU-Net unfolds $K$ iterations of this AO process into a deep cascade of $K$ stages.

### C. HSI Reconstruction Module (SR Module)

With $R$ initialized and fixed, the HSI reconstruction sub-problem is formulated as:

$$\hat{X} = \underset{x}{argmin}\frac{1}{2}\|Y - RX\|^2 + \tau\varphi(X) \quad (8)$$

To solve this, we employ the Half-Quadratic Splitting (HQS) algorithm. By introducing an auxiliary variable Z (where $Z = X$), Eq. (8) is reformulated as a constrained optimization problem:

$$\hat{X} = \underset{x}{argmin}\frac{1}{2}\|Y - RX\|^2 + \tau\varphi(Z), s.t. Z = X \quad (9)$$

Converting this to an unconstrained problem via a penalty parameter $\mu$ yields:

$$L_u(X,Z) = \frac{1}{2}\|Y - RX\|^2 + \tau\varphi(Z) + \frac{\mu}{2}\|Z - X\|^2 \quad (10)$$

This can be solved by alternately minimizing with respect to $Z$ and $X$:

$$Z_{k+1} = \underset{z}{argmin}\frac{\mu}{2}\|Z - X_{k+1}\|^2 + \tau\varphi(Z) \quad (11)$$

$$X_{k+1} = \underset{x}{argmin}\|Y - RX\|^2 + \mu\|X - Z_k\|^2 \quad (12)$$

In practice, we form this iterative process with two corresponding layers, one is the deep **Network-based Layer (Z-Net)**, and the other is the **Physics-Based layer (HSI Solver)**:

**Network-based Layer (Z-Net):** The Z-update (Eq. 11) corresponds to the proximal operator of the prior $\varphi(Z)$. We replace this operator with a learnable convolutional neural network, Z-Net, which implicitly learns data-driven priors.

**Physics-Based Solver Layer (HSI Solver)**: The X-update (Eq. 12) is a quadratic optimization problem with a unique, closed-form solution:

$$X_{k+1} = (R_k^T R_k + \mu I)^{-1}(R_k^T Y + \mu Z_k) \quad (13)$$

where $I$ is an identity matrix and $R_k$ denotes the STF used in the current stage.

### D. STF Estimation Module (STF Module)

Symmetrically, with the HSI $X$ fixed, the STF estimation s formulated as:

$$\hat{R} = \underset{R}{argmin}\|Y - RX\|^2 + \varepsilon\phi(R) \quad (14)$$

Applying HQS with an auxiliary variable $P$ (where $P = R$) and penalty parameter $\sigma$, the objective function becomes:

$$L(R) = \frac{1}{2}\|Y - RX\|^2 + \varepsilon\phi(P) + \frac{\sigma}{2}\|P - R\|^2 \quad (15)$$

This leads to two alternating sub-problems:

$$P_{k+1} = \underset{P}{argmin}\frac{\sigma}{2}\|P - R_k\|^2 + \varepsilon\phi(P) \quad (16)$$

$$R_{k+1} = \underset{R}{argmin}\|Y - RX\|^2 + \sigma\|P_{k+1} - R\|^2 \quad (17)$$

Similar to the SR module, this process is unfolded into two layers:

**Network-based Layer (P-Net):** A learnable P-Net replaces the $P$-update (Eq. 16) to capture structural priors of the spectral transformation.

**Physics-Based Solver Layer (STF-Solver):** The $R$-update (Eq. 17) is computed via its closed-form solution:

$$R_{k+1} = (X_k X_k^T + \sigma I)^{-1}(Y X_k^T + \sigma P_k) \quad (18)$$

This layer analytically updates the STF based on the current HSI estimate.

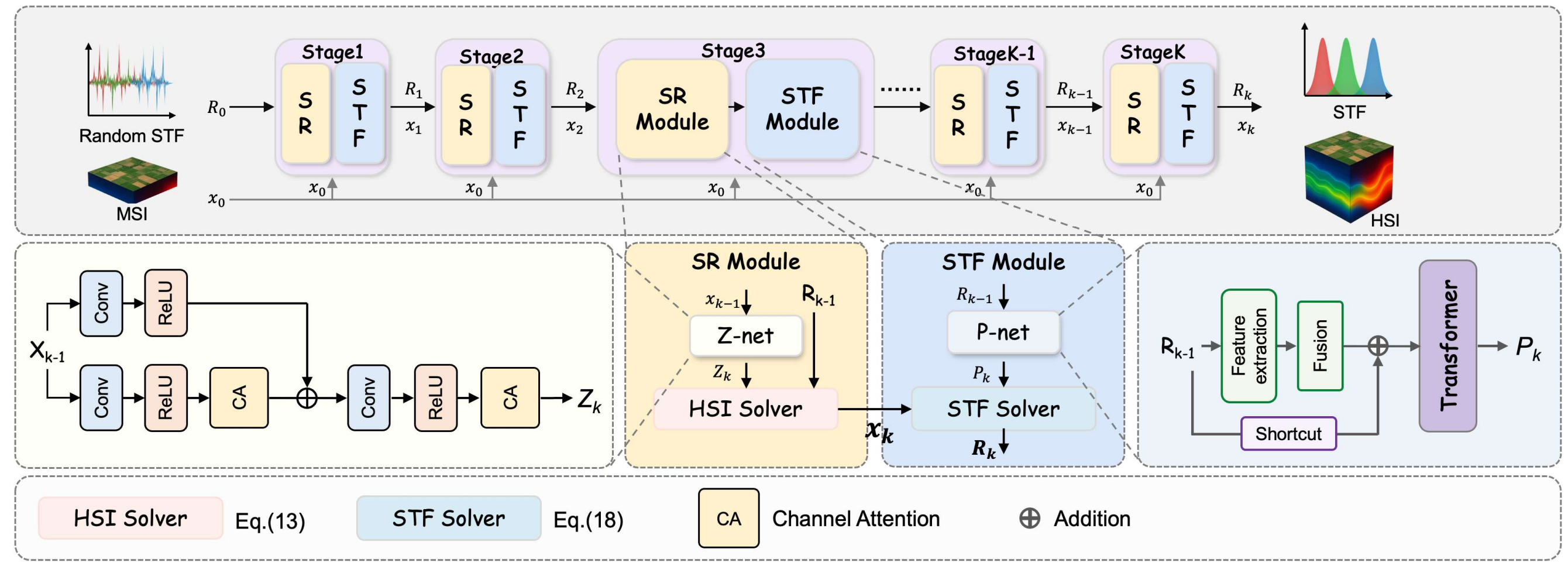

**Fig. 2.** Architecture of the overall network.

### E. Overall PGU-Net Architectures

The complete PGU-Net is formed by cascading $K$ stages, as illustrated in **Fig. 2**. Each stage sequentially contains an SR Module and an STF Module. The data flow strictly adheres to the AO algorithm: the output $X_{k+1}$ from the $k_{th}$ SR module is fed into the $k_{th}$ STF module, and the updated $R_{k+1}$ is passed to the $k + 1_{th}$ SR module. The penalty parameters $\mu$ and $\sigma$ are learned end-to-end.

1) **Z-Net**: The architecture employs a dual-branch design. The upper branch utilizes standard convolutions for global feature extraction, while the lower branch incorporates Channel Attention (CA) mechanisms to emphasize local spectral features. Features are fused via concatenation and refined through attention-enhanced convolutional layers to produce the regularized term $Z_k$.

2) **P-Net**: Taking the current estimate $R_k$ as input, P-Net utilizes a multi-branch architecture with varying kernel sizes to

extract multi-scale features. These are fused and processed by a transformer-based structure, which effectively models long-range dependencies and subtle spectral variations, yielding the regularized estimate $P_k$.

**Algorithm 1** The PGU-Net

**Input:** $\boldsymbol{Y}$

**Initialization** $\boldsymbol{R}_0$ by random initialization

**For** ***k steps*** **do:**

**Processing in SR module:**

Update $\boldsymbol{Z}$ via ***Z-Net***,

Update $\boldsymbol{X}$ via Eq. (13),

**Processing in STF module**

Update $\boldsymbol{P}$ via ***P-Net***,

Update $\boldsymbol{R}$ via Eq. (18),

**End**

**Output:** $X_k$, $R_k$.

### *F. Loss Function*

To train the network end-to-end, we employ a composite loss function balancing spatial reconstruction accuracy and spectral fidelity. The total loss $L$ is a weighted sum of the Mean Absolute Error (MAE) and the Spectral Angle Mapper (SAM) [35] :

$$L = L_{MAE} + \delta L_{SAM} \tag{19}$$

where $\delta$ is the balance parameter.  in our experiments. The MAE term, $L_{MAE}$, measures the absolute difference between the reconstructed HSI $\hat{x}_i$ and the ground truth HSI $x_i$.

$$L_{MAE} = \min_{\theta} \frac{1}{N} \sum_{i=1}^{N} \|x_i - \mathcal{F}(\hat{x}_i, \Theta)\|_1 \tag{20}$$

The SAM loss, $L_{SA}$ , measures spectral similarity by calculating the angle between corresponding spectral vectors pixel by pixel.

$$L_{SAM} = \frac{1}{N} \sum_{i=1}^{N} arccos\left(\frac{x_i \cdot \hat{x}_i}{\|x_i\|_2 \|\hat{x}_i\|_2}\right) \tag{21}$$

## III. Experimental And Results

First, we evaluate PGU-Net in controlled single-sensor simulations (Section III-B), where MSI is synthesized from HSI using a known SRF. In this setting, the ground-truth degradation operator is available, enabling quantitative evaluation of both STF estimation and HSI reconstruction.

We then evaluate PGU-Net on a real UAV cross-sensor dataset (Section III-C), where the ground-truth STF is unknown. In this case, reconstruction quality is assessed by comparing the recovered HSI with the reference HSI captured by the hyperspectral sensor.

### *A. Datasets and Implementation Details*

***Datasets:*** We use two datasets to evaluate model performance under (i) simulated single-sensor settings and (ii) real cross-sensor settings.

**1). Simulated Single-Sensor Datasets (controlled validation):** To assess the blind estimation and reconstruction capability of PGU-Net under a controlled setting and to enable fair comparisons with SSR methods commonly used in the computer vision community, we adopt two benchmark datasets: CAVE[1][36] (32 HSIs, 512×512×31, 400–700 nm) and NTIRE 2022 [37] (1000 HSIs, 482×512×31). We randomly split the datasets into 20/12 (CAVE) and 300/150 (NTIRE 2022) training/testing samples, respectively. For both datasets, MSI is synthesized from HSI using a known SRF. Following common practice, we use the Nikon D700[2] SRF as the default setting. This enables quantitative evaluation of both HSI reconstruction and STF estimation. Besides Nikon D700, we further test three additional SRFs that are widely used in remote sensing: Landsat-7 ETM+, Sentinel-2 MSI, and randomly generated Gaussian SRFs.

**2). Real-World Cross-Sensor Dataset (practical evaluation).** We use a co-registered Headwall[3]-DJI[4] UAV dataset acquired over a lychee orchard in Shenzhen. The dataset contains a Headwall Nano hyperspectral image (400–1000 nm, 144 bands used) and a DJI P4 Multispectral image (four bands centered at 450, 560, 650, and 730 nm). Both modalities are cropped to 7790×6032 pixels. To reduce spatial leakage between training and testing, we split spatially disjoint regions for training and testing (70% on the left for training and the remaining 30% on the right for testing), as illustrated in **Fig. 3**.

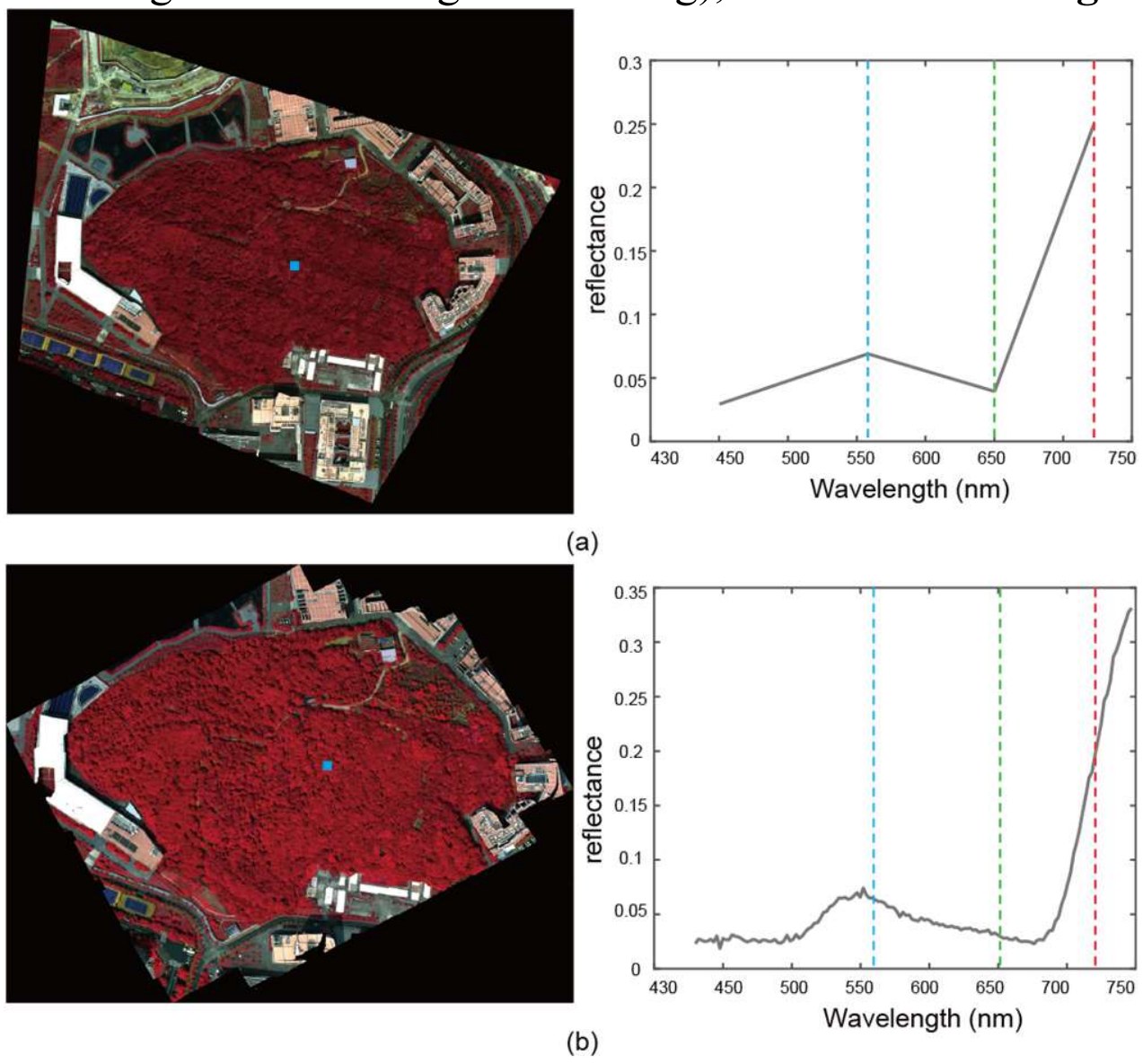

**Fig. 3.** Headwall–DJI cross-sensor UAV dataset. Left: false-color composites (bands indicated by dashed lines) with a selected tree pixel (blue). Right: corresponding spectra from DJI MSI and Headwall HSI at the marked pixel.

***Compared Methods:*** In this study, we compare PGU-Net with representative SSR methods, including HSCNN+ [38], AWAN [39], MST++ [40] and SSR [26]. HSCNN+ and AWAN are CNN-based methods, MST++ is transformer-based, and SSR incorporates a physical degradation model with deep learning. All compared methods are retrained on the same training split with identical patch extraction and data normalization for fair comparison.

[1] http://www.cs.columbia.edu/CAVE/databases/multispectral/
[2] https://maxmax.com/spectral_response.htm
[3] https://headwallphotonics.com/
[4] https://www.dji.com/cn/support/product/p4-multispectral

***Implementation***: PGU-Net was implemented in PyTorch [41] and trained on an Intel i7-13700K CPU with an NVIDIA RTX 4080 GPU. Unless stated otherwise, we used $K = 6$ stages. Z-Net employs a 64-channel convolution with LeakyReLU and a channel-attention branch. P-Net uses two 1D convolutions (16 and 32 channels) followed by a Transformer (positional encoding, 4-head attention, FFN hidden size 64). The HQS penalty parameters $\mu$ and $\sigma$ were initialized to 0.01 and learned during training. Training samples were generated by cropping paired HSI/MSI patches of size 64×64 with batch size 64. Multi-scale convolutions (1×1, 3×3, and 5×5) are fused with learnable weights in the attention module. We trained the model using Adam to minimize MAE+SAM with weights [1, 0.1].

***Quality Metrics:*** For performance evaluation, three quantitative Perceptual Quality Indicators (PQI) were used: Root Mean Square Error (RMSE), the Peak Signal-to-Noise Ratio (PSNR), and the Structural Similarity Index Measure (SSIM).

### *B. Performance on Simulated Single-Sensor SSR scenario*

We first evaluate PGU-Net under simulated single-sensor settings, where MSI is synthesized from HSI using a known SRF. This setting provides ground-truth supervision for evaluating both STF estimation and HSI reconstruction.

In the simulated single-sensor scenario, the estimated STF $\hat{R}$ can be directly validated against the known $SRF_{msi}$. In the following experiments, we evaluate STF recovery and HSI reconstruction on both CAVE and NTIRE 2022 under four SRFs: Nikon D700, Landsat-7 ETM+, randomly generated Gaussian SRFs, and Sentinel-2 MSI.

1) ***Accuracy of STF Recovery:*** **Fig. 4** compares the STF estimated by PGU-Net (i.e., the learned R) with the ground-truth Nikon D700 SRF on CAVE and NTIRE 2022. The estimated curves closely match the ground truth in terms of peak locations and overall spectral shapes. The largest deviations are observed in the red band, which may be related to band-dependent signal-to-noise ratios.

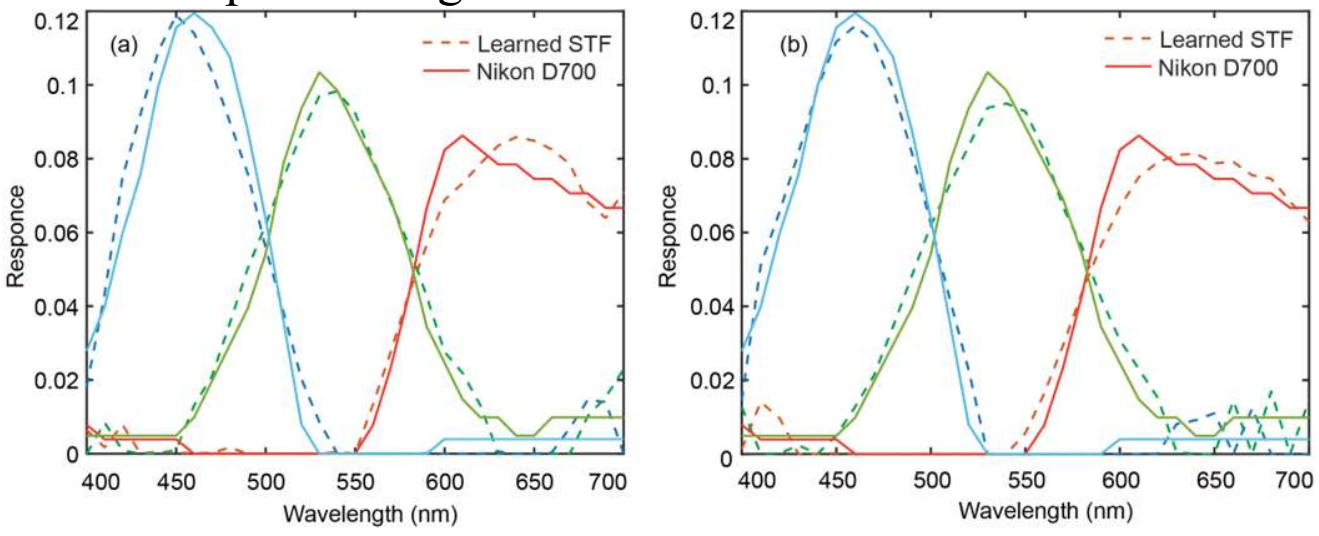

**Fig. 4.** A comparison of the SRFs recovered by the model (dashed line) with the ground truth (solid line); (a) CAVE dataset; (b) NTIRE 2022 dataset.

To quantify the impact of these deviations, we use the estimated STF to degrade HSI and obtain a reconstructed MSI, which is then compared with the input MSI synthesized using the ground-truth SRF. Table I reports the quantitative results. The reconstructed MSIs are nearly identical to the ground-truth MSIs, with very low RMSE and high PSNR/SSIM.

TABLE I

EVALUATION OF SPECTRAL DEGRADATION EFFECTIVENESS.

| | RMSE | PSNR | SSIM |
|---|---|---|---|
| CAVE | 0.0017 | 55.31 | 0.997 |
| NTIRE 2022 | 0.0013 | 54.98 | 0.998 |

**Fig. 5.** Comparison of MSI reconstruction by STF predicted by PGU-Net on **CAVE dataset**. (1) MSI recovered by Nikon D700 SRF; (2) MSI recovered by STFs predicted by PGU-Net; (3-5) RMSE error maps between (1) and (2) in red, green, and blue channels. (a-e) different samples named “Jellybeans”, “oil painting”, “paints”, “photo and face”, “pompoms”.

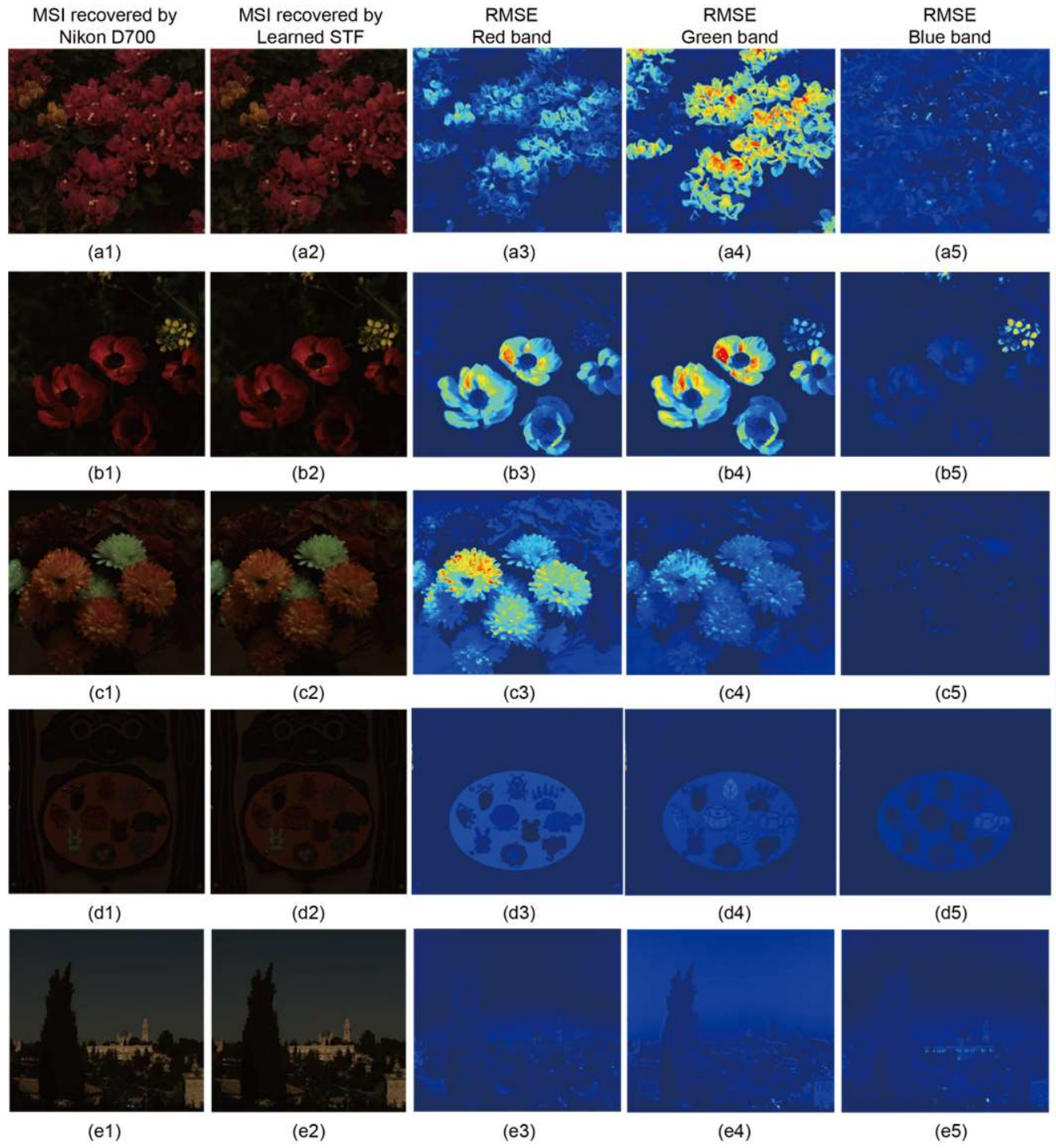

**Fig. 6.** Comparison of RGB reconstruction on NTIRE 2022: (1) MSI with ground-truth SRF; (2) MSI with SRFs recovered by model; (3-5) Error maps in red, green, and blue bands. (a-e) different samples named "ARAD_1K_0636", "ARAD_1K_0671", "ARAD_1K_0621", "ARAD_1K_0945", "ARAD_1K_0948".

We further evaluate STF estimation under three additional SRFs: Landsat-7 ETM+, randomly generated Gaussian SRFs, and Sentinel-2 MSI. **Fig. 7** shows that PGU-Net accurately recovers the peak positions and spectral support of each band across different SRFs. Table II reports the corresponding HSI reconstruction performance on CAVE. These results indicate that PGU-Net can reliably estimate the STF from paired data, which is critical for real cross-sensor settings where the STF is unknown.

2) ***Hyperspectral Image Reconstruction Quality Assessment:*** After validating STF estimation, we evaluate HSI reconstruction against the ground-truth HSI. **Fig. 8** provides spectral comparisons on the CAVE dataset, where the reconstructed spectra closely match the ground truth at selected pixels, preserving absorption valleys and reflectance peaks.

We further compare PGU-Net with HSCNN+, AWAN, MST++, and SSR. Table III summarizes the quantitative results on CAVE and NTIRE 2022. PGU-Net consistently achieves the best performance across all metrics, which is also reflected in the RMSE maps in **Fig. 9**.

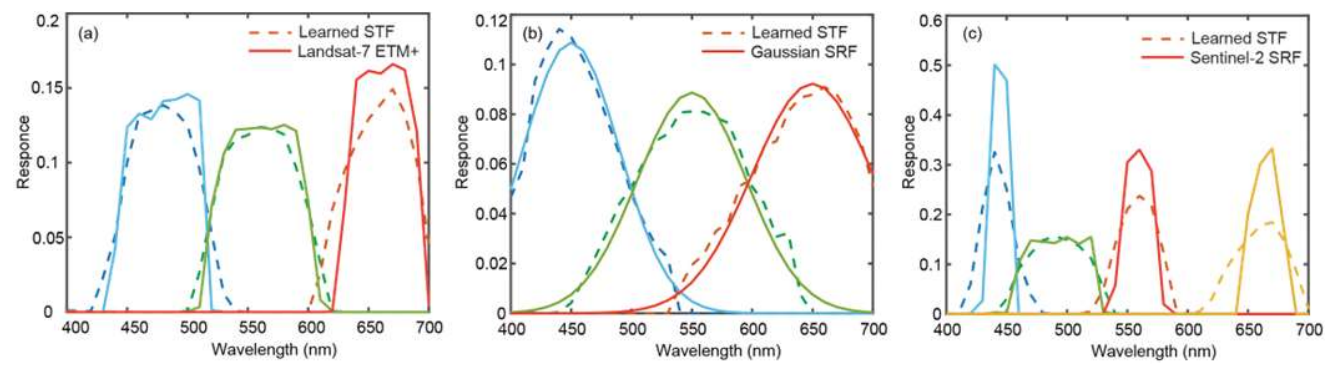

**Fig. 7.** Recovered STFs (dashed lines) versus ground-truth SRFs (solid lines) for three given sensor types. (a) Landsat-7 ETM+. (b) Randomly generated Gaussian SRF. (c) Sentinel-2 MSI SRF.

TABLE II
HSI RECONSTRUCTION PERFORMANCE (MEAN ± STD. DEV.) ON CAVE DATASET WITH FOUR GIVEN SRFS.

| | RMSE | PSNR | SSIM |
|---|---|---|---|
| Nikon D700 | 0.0163± 0.0002 | 36.32 ± 0.11 | 0.9818 ± 0.0001 |
| Landsat-7 | 0.0176 ± 0.0001 | 35.70 ± 0.04 | 0.9787 ± 0.0001 |
| Gaussian SRF | 0.0159 ± 0.0002 | 36.54 ± 0.09 | 0.9817 ± 0.0004 |
| Sentinel-2 MSI | 0.0168 ± 0.0002 | 35.98 ± 0.11 | 0.9817 ± 0.0002 |

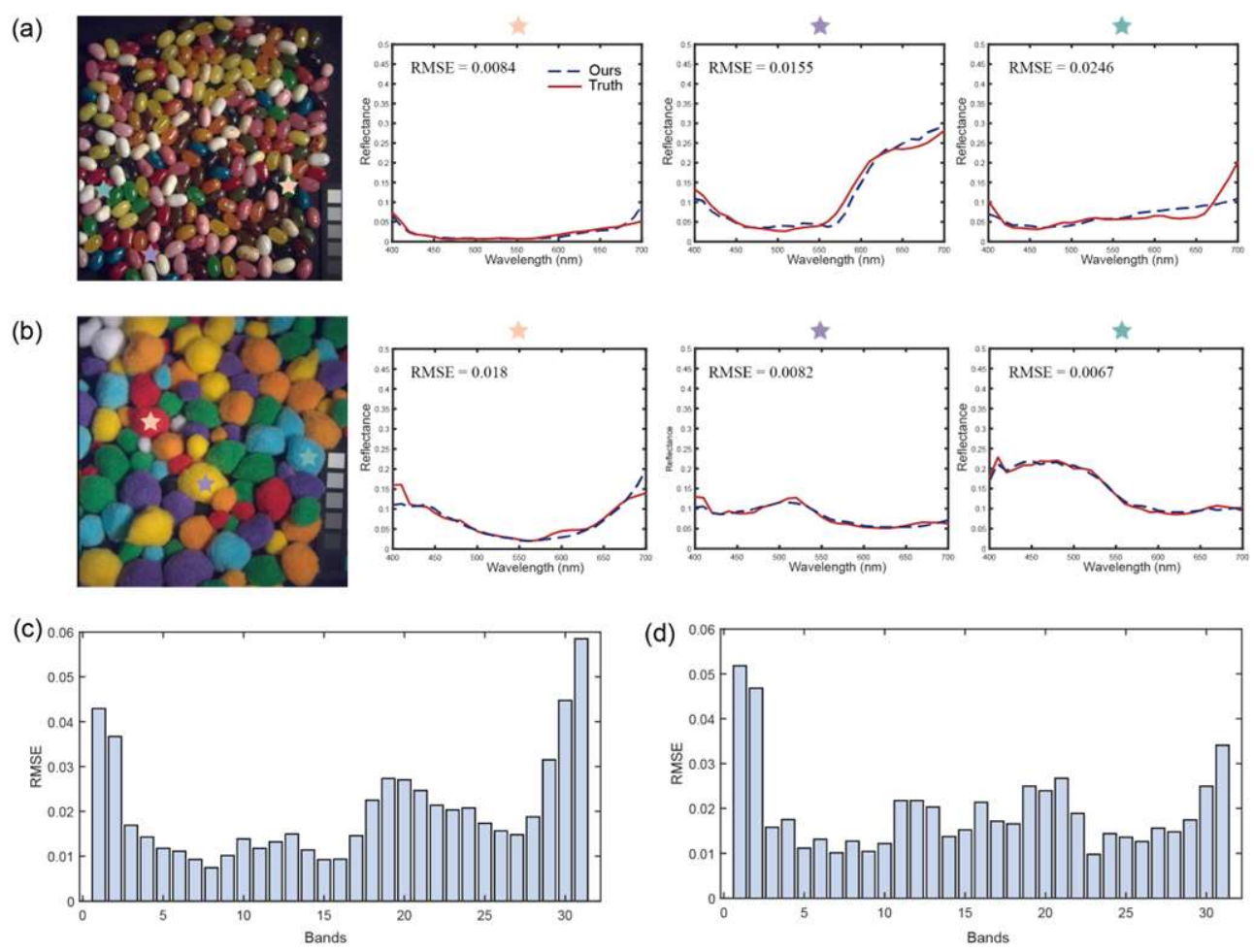

**Fig. 8**. Comparison of the super-resolved HSI and ground truth on the CAVE dataset. (a) Image named "jellybeans"; (b) image named "pompoms"; (c) histogram of RMSE over 31 bands in "jellybeans"; (d) histogram of RMSE over 31 bands in "pompoms".

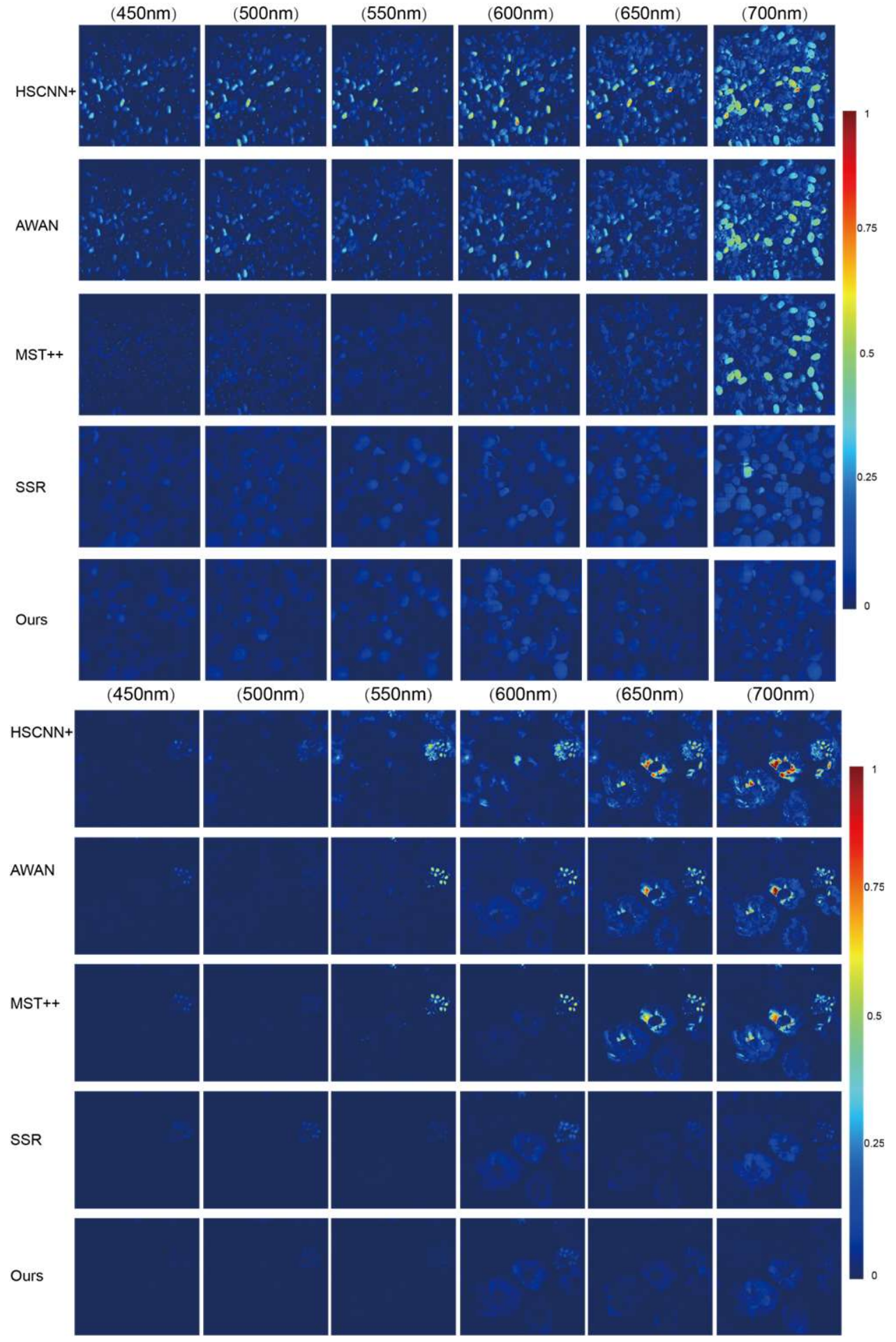

**Fig. 9.** RMSE error map of different SSR methods comparison on (Up) the CAVE dataset "pompoms" ;(Down) NTIRE 2022 "ARAD_1K_0671".

TABLE III
QUANTITATIVE COMPARISON OF DIFFERENT METHODS ON THE CAVE AND NTIRE 2022 DATASET

| | CAVE | | | NTIRE 2022 | | |
|---|---|---|---|---|---|---|
| | RMSE | PSNR | SSIM | RMSE | PSNR | SSIM |
| HSCNN+ | 0.0494 | 26.18 | 0.9389 | 0.0588 | 26.36 | 0.9438 |
| AWAN | 0.0401 | 27.93 | 0.9550 | 0.0317 | 32.12 | 0.9646 |
| MST++ | 0.0281 | 31.02 | 0.9709 | 0.0247 | 34.32 | 0.9805 |
| SSR | 0.0194 | 34.95 | 0.9765 | 0.0110 | 39.16 | 0.9921 |
| **PGU-Net** | 0.0164 | 36.28 | 0.9816 | 0.0086 | 41.35 | 0.9926 |

### *C. Performance on Real-World Cross-Sensor Scenario*

Having established PGU-Net's superior performance in a controlled environment, we now assess its effectiveness on the Headwall-DJI UAV dataset, which represents a true blind, cross-sensor challenge.

First, we visualized the cross-sensor STF recovered by PGU-Net, which represents the learned transformation from the Headwall HSI to the DJI MSI space. As shown in **Fig. 10**, PGU-Net learns a physically plausible STF between sensors, where the response peaks align with the DJI band centers (approximately 450, 560, 650, and 730 nm). This is expected because the MSI provides observations only in these four spectral bands, which primarily constrain the STF in the corresponding wavelength regions.

This learned STF is then used to guide the HSI reconstruction. **Fig. 11** presents the qualitative HSI reconstruction results. In **Fig. 11**(a), we show the experimental area, and in **Fig. 11**(b), we compare the reconstructed spectral curves with the ground truth Headwall spectra for three representative points. The plots show that PGU-Net reconstructs characteristic spectral features across diverse land-cover types, including forest (b1), roads (b4, b6), roofs (b3, b7, b8), and mixed regions (b2, b5). We further compare PGU-Net with HSCNN+, AWAN, and MST++ in this cross-sensor setting. The physics-based SSR method [31] is excluded because it requires a known SRF. As shown in **Fig. 12**, PGU-Net achieves the lowest reconstruction error, demonstrating robustness under truly blind conditions.

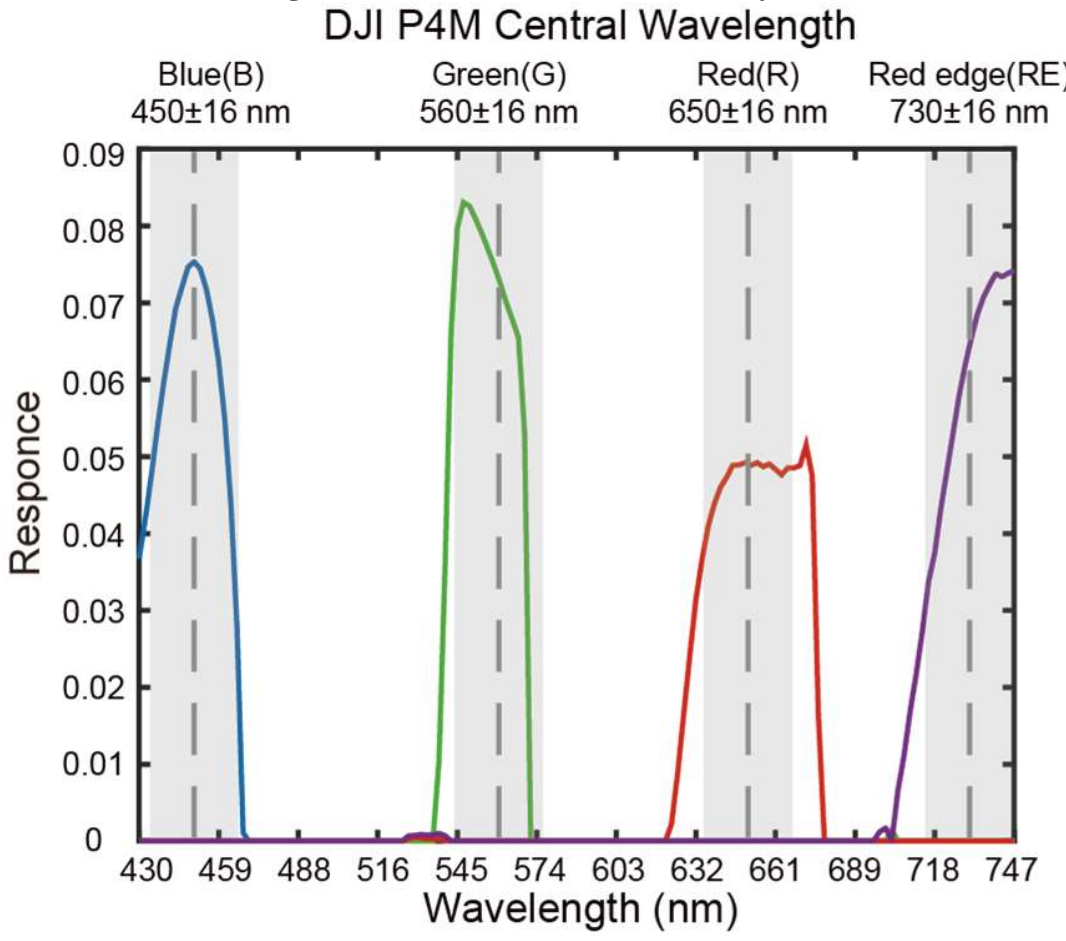

**Fig. 10.** Cross-sensor STF between DJI P4M and Headwall Nano learn by model. The dashed line is the center wavelength of the DJI P4M multispectral sensor.

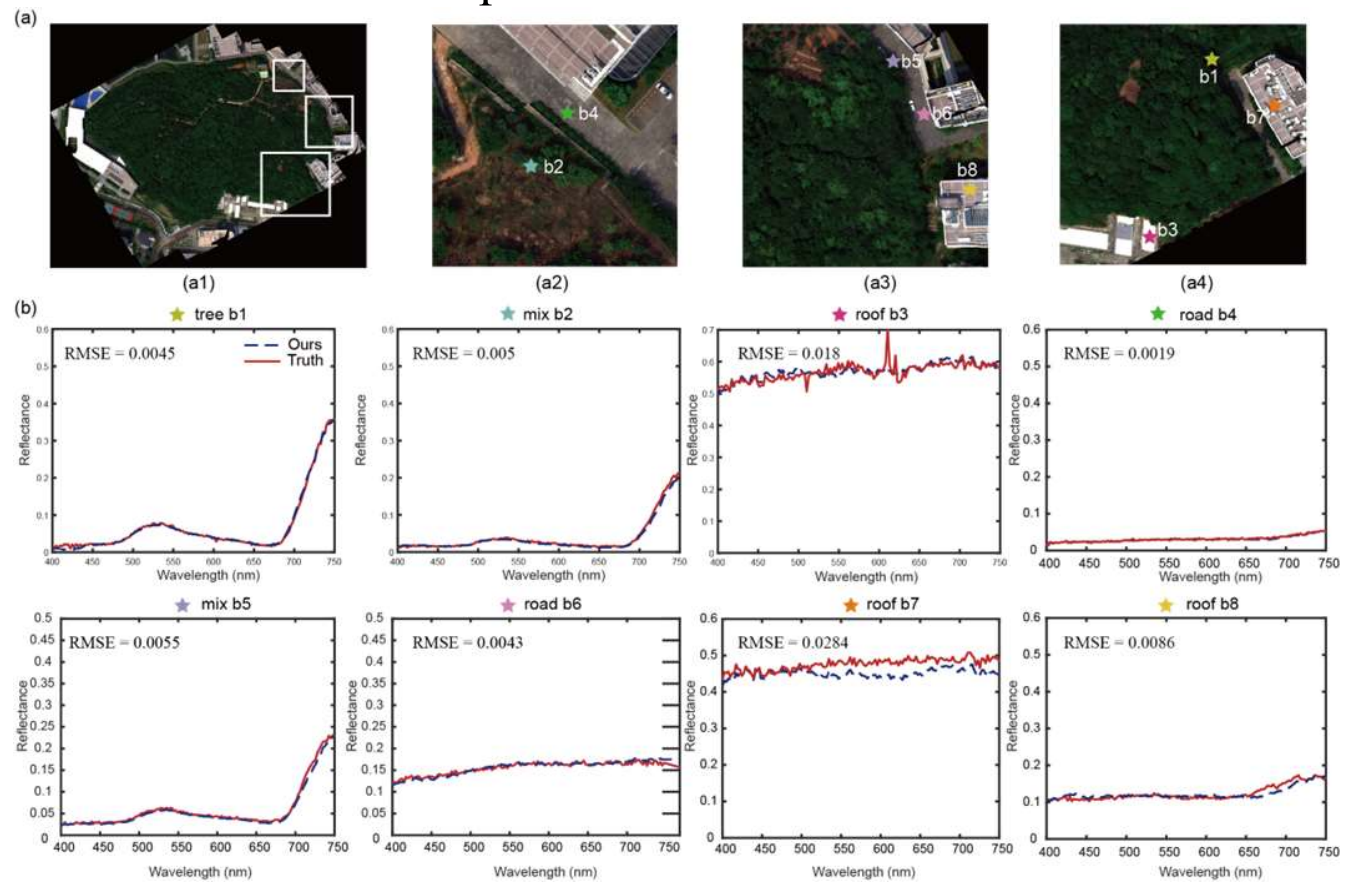

**Fig. 11.** Comparison between super-resolved HSI and ground truth for diverse land cover types, including forest (b1), roads (b4, b6), roof (b3, b7 and b8) and mixed (b2, b5) region.

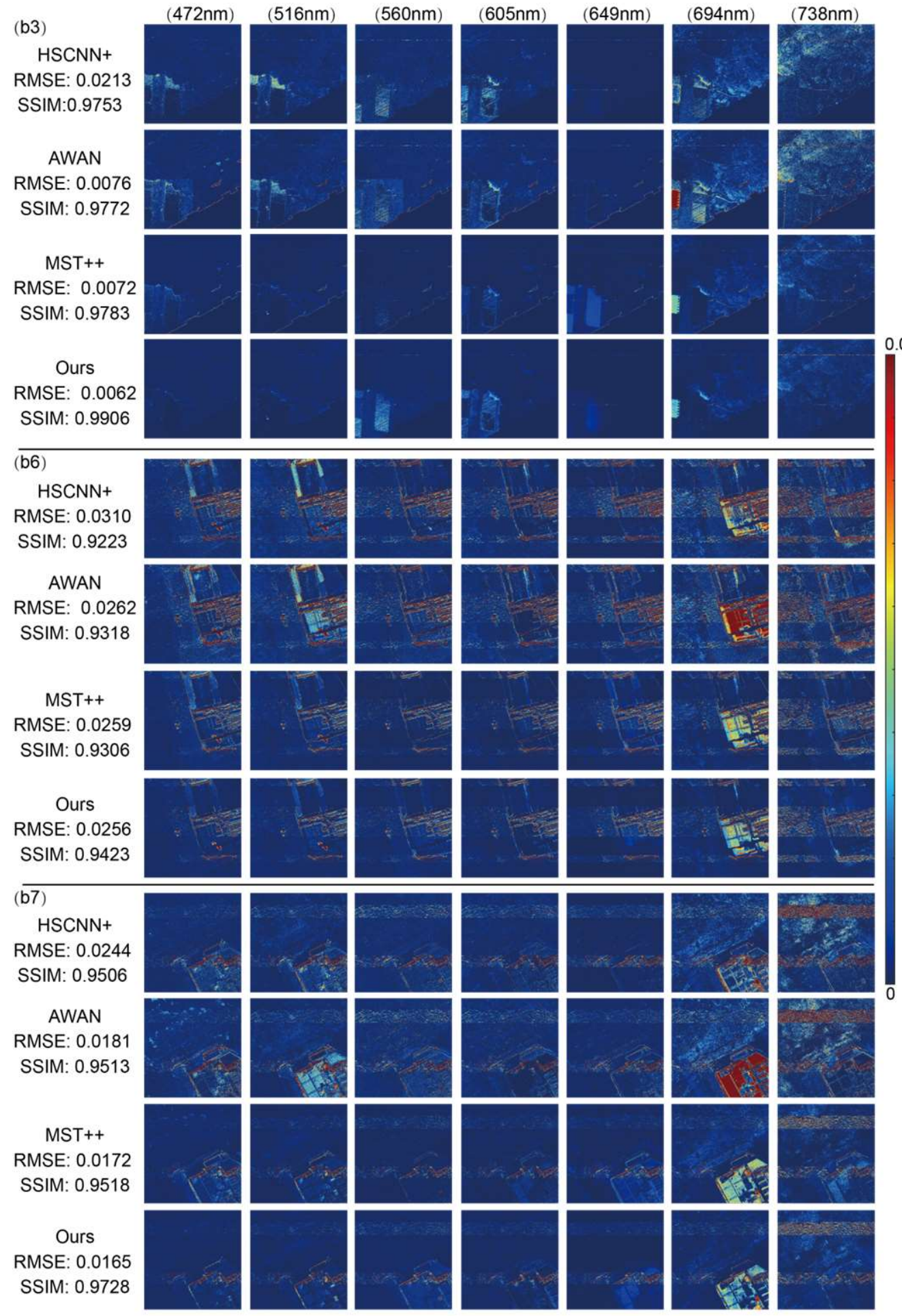

**Fig. 12.** RMSE maps of super-resolved HSI results from roof region (b3, b7) and road region (b6) in Fig. 11(a).

## IV LAND-COVER DEPENDENCY OF THE STF AND ABLATION STUDY

In this section, we use PGU-Net as an analysis tool to examine whether the estimated STF varies with land-cover types (Section IV-A). We also conduct ablation and stability studies to evaluate key components and training robustness (Section IV-B).

### *A. Investigation into the Land-Cover Dependency of the STF*

The validated PGU-Net can serve as a computational tool to examine a fundamental question in SSR: is the optimal STF static across a scene? To explore this, we trained specialist models on distinct, relatively pure land-cover patches (tree canopy and building) from our UAV dataset.

According to Eq. (4), the STF between two sensors is jointly determined by SRF-related terms and SMF-related terms. The SRF-related terms are sensor-specific and remain fixed for a given instrument, whereas the SMF-related terms can vary with scene content and acquisition geometry.

**Fig. 13** compares STFs estimated from (a) all pixels, (b) tree pixels, and (c) building pixels. The observed differences suggest that the effective STF may depend on scene content, which is consistent with the scene-dependent nature of spectral mixing. We emphasize that this analysis is exploratory; a more rigorous statistical validation (e.g., repeated trials, controlled sampling, and uncertainty quantification) will be investigated in future work.

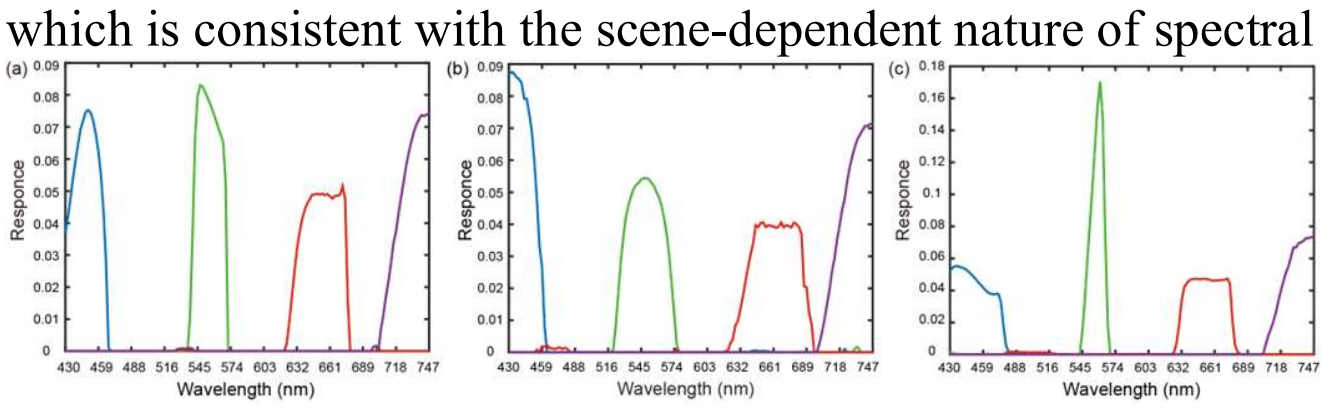

**Fig. 13.** (a) STFs recovered by all pixels; (b) STFs recovered by only tree pixels; (c) STFs recovered by only building pixels.

These observations motivate the use of a learnable STF in cross-sensor SSR and indicate that the estimated STF may provide a useful cue for analyzing cross-sensor spectral transformation under different land-cover compositions. For a fixed sensor pair, this perspective naturally separates STF-related factors into sensor-dependent components (e.g., SRF/PSF) and scene-dependent components associated with spectral mixing, which may vary across land-cover types.

### *B. Model Analysis and Robustness*

Finally, we conduct a series of experiments to dissect the model's architecture, validate its stability, and demonstrate the robustness of the overall framework with the ablation study.

**Ablation Study:** To justify our network design and quantify the contribution of its key components, we conducted an ablation study on the CAVE dataset. The results are summarized in Table IV. When the STF estimation module is disabled (w/o STF Module), the performance degrades most severely, highlighting the critical importance of the blind estimation capability. Removing the attention mechanisms in the P-Network (w/o SA) and Z-Network (w/o CA) also leads to a noticeable drop in accuracy. Encouragingly, our full PGU-Net significantly closes the performance gap to the non-blind upper bound (w/ Ground-Truth STF), confirming the effectiveness of our joint optimization approach.

Table IV
ABLATION STUDY ON CAVE DATASET

| | RMSE | PSNR | SSIM |
|---|---|---|---|
| PGU-Net | 0.0164 | 36.28 | 0.9816 |
| w/o CA | 0.0167 | 36.08 | 0.9809 |
| w/o SA | 0.0165 | 36.26 | 0.9811 |
| w/o STF Module | 0.0195 | 34.87 | 0.9757 |

**Stability and Sensitivity to Initialization:** To verify that the model's performance is not dependent on a fortuitous random initialization, we trained the entire framework five times on the CAVE dataset (using the Nikon D700 SRF), each with a different random seed. Table V summarizes the statistics of the final HSI reconstruction metrics, showing extremely low coefficients of variation and confirming a highly stable and reproducible training outcome. This stability is visually corroborated in **Fig. 14**, which shows that the STFs recovered from the five independent runs converge to a nearly identical solution, tightly bracketing the ground truth.

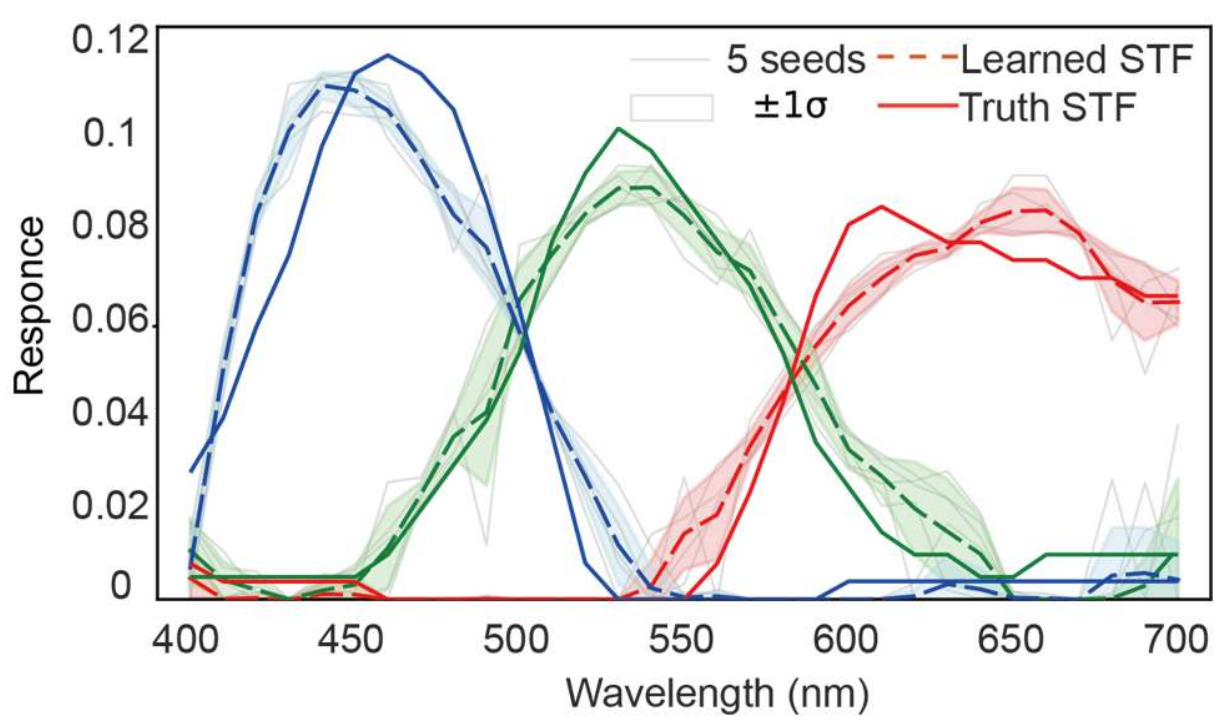

**Fig. 14.** Consistency of the recovered STF across five random seeds. Gray curves: individual runs; shaded band: mean ±1 std; red curve: ground-truth SRF (Nikon D700).

Table V
FIVE INDEPENDENT RUNS WITH DIFFERENT RANDOM SEEDS (CAVE DATASET).

| Experiments | RMSE | PSNR | SSIM |
|---|---|---|---|
| Seed 42 | 0.0164 | 36.28 | 0.9816 |
| Seed 123 | 0.0170 | 36.00 | 0.9797 |
| Seed 1024 | 0.0167 | 36.18 | 0.9806 |
| Seed 2025 | 0.0166 | 36.14 | 0.9809 |
| Seed 5555 | 0.0170 | 35.99 | 0.9791 |
| Mean | 0.0167 | 36.12 | 0.9804 |
| Std. | 0.0002 | 0.11 | 0.0009 |

## V. Conclusion

This paper presented PGU-Net, a physics-guided deep unfolding framework for blind cross-sensor spectral super-resolution. By formulating SSR as a joint estimation problem of the target HSI and an explicit spectral transformation matrix, PGU-Net integrates interpretable model-based solvers with learnable proximal networks in an end-to-end trainable architecture. Experiments on simulated benchmarks demonstrate accurate recovery of the degradation operator and improved reconstruction performance over representative SSR methods. Evaluations on a real UAV cross-sensor dataset further validate effectiveness under truly blind conditions and suggest that the estimated STF may vary with land-cover categories. Future work will focus on strengthening the statistical validation of STF variability, improving robustness to registration and illumination changes, and extending the framework to broader cross-scene generalization.

## Acknowledgment

This research was supported by the National Key Research and Development Program of China (Project Nos. 2023YFF1303605), Guangdong Basic and Applied Basic Research Foundation(2024A1515030014) and the Fundamental Research Foundation of Shenzhen Technology and Innovation Council (Project Nos. KCXST20221021111611029, JCYJ20220818101617037).

## References

[1] M. B. Stuart, A. J. S. McGonigle, and J. R. Willmott, "Hyperspectral Imaging in Environmental Monitoring: A Review of Recent Developments and Technological Advances in Compact Field Deployable Systems," *Sensors-basel*, vol. 19, no. 14, Art. no. 14, Jan. 2019, doi: 10.3390/s19143071.

[2] F. Mena *et al.*, "Adaptive fusion of multi-modal remote sensing data for optimal sub-field crop yield prediction," *Remote Sens Environ*, vol. 318, p. 114547, Mar. 2025, doi: 10.1016/j.rse.2024.114547.

[3] J. Joiner *et al.*, "The seasonal cycle of satellite chlorophyll fluorescence observations and its relationship to vegetation phenology and ecosystem atmosphere carbon exchange," *Remote Sensing of Environment*, vol. 152, pp. 375–391, Sept. 2014, doi: 10.1016/j.rse.2014.06.022.

[4] L. Deng, Z. Mao, X. Li, Z. Hu, F. Duan, and Y. Yan, "UAV-based multispectral remote sensing for precision agriculture: A comparison between different cameras," *Isprs J Photogramm*, vol. 146, pp. 124–136, Dec. 2018, doi: 10.1016/j.isprsjprs.2018.09.008.

[5] J. He *et al.*, "Spectral super-resolution meets deep learning: Achievements and challenges," *Inform Fusion*, vol. 97, p. 101812, Sept. 2023, doi: 10.1016/j.inffus.2023.101812.

[6] Q. Wang, Y. Tang, and P. M. Atkinson, "The effect of the point spread function on downscaling continua," *ISPRS Journal of Photogrammetry and Remote Sensing*, vol. 168, pp. 251–267, Oct. 2020, doi: 10.1016/j.isprsjprs.2020.08.016.

[7] T. Okamoto and I. Yamaguchi, "Simultaneous acquisition of spectral image information," *Opt. Lett.*, vol. 16, no. 16, p. 1277, Aug. 1991, doi: 10.1364/OL.16.001277.

[8] F. Agahian, S. A. Amirshahi, and S. H. Amirshahi, "Reconstruction of reflectance spectra using weighted principal component analysis," *Color Research & Application*, vol. 33, no. 5, pp. 360–371, Oct. 2008, doi: 10.1002/col.20431.

[9] N. Eslahi, S. H. Amirshahi, and F. Agahian, "Recovery of spectral data using weighted canonical correlation regression," *OPT REV*, vol. 16, no. 3, pp. 296–303, May 2009, doi: 10.1007/s10043-009-0055-y.

[10] V. Cheung, C. Li, J. Hardeberg, D. Connah, and S. Westland, "Characterization of trichromatic color cameras by using a new multispectral imaging technique," *J. Opt. Soc. Am. A*, vol. 22, no. 7, p. 1231, July 2005, doi: 10.1364/JOSAA.22.001231.

[11] B. Arad and O. Ben-Shahar, "Sparse recovery of hyperspectral signal from natural RGB images," in *Computer Vision – ECCV 2016*, B. Leibe, J. Matas, N. Sebe, and M. Welling, Eds., Cham: Springer International Publishing, 2016, pp. 19–34. doi: 10.1007/978-3-319-46478-7_2.

[12] A. Robles-Kelly, "Single image spectral reconstruction for multimedia applications," in *Proceedings of the 23rd ACM international conference on Multimedia*, in MM '15. New York, NY, USA: Association for Computing Machinery, Oct. 2015, pp. 251–260. doi: 10.1145/2733373.2806223.

[13] N. Akhtar and A. Mian, "Hyperspectral Recovery from RGB Images using Gaussian Processes," *Ieee T Pattern Anal*, vol. 42, no. 1, pp. 100–113, Jan. 2020, doi: 10.1109/TPAMI.2018.2873729.

[14] Y. Jia *et al.*, "From RGB to spectrum for natural scenes via manifold-based mapping," in *2017 IEEE International Conference on Computer Vision (ICCV)*, Oct. 2017, pp. 4715–4723. doi: 10.1109/ICCV.2017.504.

[15] S. Galliani, C. Lanaras, D. Marmanis, A. N. for S. C.-A. S. I. R. from R. Baltsavias, and K. Schindler, "Learned Spectral Super-Resolution," Mar. 28, 2017, *arXiv*: arXiv:1703.09470. doi: 10.48550/arXiv.1703.09470.

[16] J. Li *et al.*, "Hybrid 2-D–3-D Deep Residual Attentional Network With Structure Tensor Constraints for Spectral Super-Resolution of RGB Images," *Ieee T Geosci Remote*, vol. 59, no. 3, pp. 2321–2335, Mar. 2021, doi: 10.1109/TGRS.2020.3004934.

[17] S. Paul and D. Nagesh Kumar, "Transformation of multispectral data to quasi-hyperspectral data using convolutional neural network regression," *IEEE Transactions on Geoscience and Remote Sensing*, vol. 59, no. 4, pp. 3352–3368, Apr. 2021, doi: 10.1109/TGRS.2020.3009290.

[18] T. Li and Y. Gu, "Progressive Spatial–Spectral Joint Network for Hyperspectral Image Reconstruction," *Semisupervised Spectral Degradation Constrained Network for Spectral Super-Resolution*, vol. 60, pp. 1–14, 2022, doi: 10.1109/TGRS.2021.3079969.

[19] X. Zheng, W. Chen, and X. Lu, "Spectral super-resolution of multispectral images using spatial–spectral residual attention network," *IEEE Transactions on Geoscience and Remote Sensing*, vol. 60, pp. 1–14, 2022, doi: 10.1109/TGRS.2021.3104476.

[20] A. Alvarez-Gila, J. Van De Weijer, and E. Garrote, "Adversarial Networks for Spatial Context-Aware Spectral Image Reconstruction from RGB," in *2017 IEEE International Conference on Computer Vision Workshops (ICCVW)*, Oct. 2017, pp. 480–490. doi: 10.1109/ICCVW.2017.64.

[21] Y. Cai *et al.*, "Mask-guided Spectral-wise Transformer for Efficient Hyperspectral Image Reconstruction," Mar. 21, 2022, *arXiv*: arXiv:2111.07910. doi: 10.48550/arXiv.2111.07910.

[22] W. Dong, S. Liu, S. Xiao, J. Qu, and Y. Li, "ISPDiff: Interpretable Scale-Propelled Diffusion Model for Hyperspectral Image Super-Resolution," *Ieee T Geosci Remote*, vol. 62, pp. 1–14, 2024, doi: 10.1109/TGRS.2024.3407967.

[23] Y. Xiao, Q. Yuan, K. Jiang, J. He, X. Jin, and L. Zhang, "EDiffSR: An Efficient Diffusion Probabilistic Model for Remote Sensing Image Super-Resolution," *Ieee T Geosci Remote*, vol. 62, pp. 1–14, 2024, doi: 10.1109/TGRS.2023.3341437.

[24] H. Shen, M. Jiang, J. Li, C. Zhou, Q. Yuan, and L. Zhang, "Coupling model- and data-driven methods for remote sensing image restoration and fusion: Improving physical interpretability," *IEEE Geoscience and Remote Sensing Magazine*, vol. 10, no. 2, pp. 231–249, June 2022, doi: 10.1109/MGRS.2021.3135954.

[25] J. He, J. Li, Q. Yuan, H. Shen, and L. Zhang, "Spectral Response Function-Guided Deep Optimization-Driven Network for Spectral Super-Resolution," *Ieee T Neur Net Lear*, vol. 33, no. 9, pp. 4213–4227, Sept. 2022, doi: 10.1109/TNNLS.2021.3056181.

[26] R. Dian, T. Shan, W. He, and H. Liu, "Spectral Super-Resolution via Model-Guided Cross-Fusion Network," *Ieee T Neur Net Lear*, pp. 1–12, 2023, doi: 10.1109/TNNLS.2023.3238506.

[27] B. Arad and O. Ben-Shahar, "Filter Selection for Hyperspectral Estimation," in *2017 IEEE International Conference on Computer Vision (ICCV)*, Oct. 2017, pp. 3172–3180. doi: 10.1109/ICCV.2017.342.

[28] Y. Fu, T. Zhang, Y. Zheng, D. Zhang, and H. Huang, "Joint Camera Spectral Response Selection and Hyperspectral Image Recovery," *Ieee T Pattern Anal*, vol. 44, no. 1, pp. 256–272, Jan. 2022, doi: 10.1109/TPAMI.2020.3009999.

[29] E. Martínez, S. Castro, J. Bacca, and H. Arguello, "Efficient transfer learning for spectral image reconstruction from RGB images," in *2020 IEEE Colombian Conference on Applications of Computational Intelligence (IEEE ColCACI 2020)*, Aug. 2020, pp. 1–6. doi: 10.1109/ColCACI50549.2020.9247895.

[30] T. Li and Y. Gu, "Progressive Spatial–Spectral Joint Network for Hyperspectral Image Reconstruction," *IEEE Transactions on Geoscience and Remote Sensing*, vol. 60, pp. 1–14, 2022, doi: 10.1109/TGRS.2021.3079969.

[31] Z. Zhen, S. Chen, T. Yin, and J.-P. Gastellu-Etchegorry, "Globally quantitative analysis of the impact of atmosphere and spectral response function on 2-band enhanced vegetation index (EVI2) over sentinel-2 and landsat-8," *ISPRS Journal of Photogrammetry and Remote Sensing*, vol. 205, pp. 206–226, Nov. 2023, doi: 10.1016/j.isprsjprs.2023.09.024.

[32] J. Xie, L. Fang, C. Wu, F. Xie, and J. Chanussot, "Blind Spectral Super-Resolution by Estimating Spectral Degradation Between Unpaired Images," *IEEE Transactions on Geoscience and Remote Sensing*, vol. 62, pp. 1–14, 2024, doi: 10.1109/TGRS.2024.3387857.

[33] Y. Qi, M. Lou, Y. Liu, L. Li, Z. Yang, and W. Nie, "Advancing image super-resolution techniques in remote sensing: A comprehensive survey," *ISPRS Journal of Photogrammetry and Remote Sensing*, vol. 231, pp. 68–100, Jan. 2026, doi: 10.1016/j.isprsjprs.2025.10.024.

[34] L. Liu, W. Li, Z. Shi, and Z. Zou, "Physics-informed hyperspectral remote sensing image synthesis with deep conditional generative adversarial networks," *IEEE Transactions on Geoscience and Remote Sensing*, vol. 60, pp. 1–15, 2022, doi: 10.1109/TGRS.2022.3173532.

[35] Y. Li, L. Zhang, C. Dingl, W. Wei, and Y. Zhang, "Single Hyperspectral Image Super-Resolution with Grouped Deep Recursive Residual Network," in *2018 IEEE Fourth International Conference on Multimedia Big Data (BigMM)*, Sept. 2018, pp. 1–4. doi: 10.1109/BigMM.2018.8499097.

[36] F. Yasuma, T. Mitsunaga, D. Iso, and S. K. Nayar, "Generalized Assorted Pixel Camera: Postcapture Control of Resolution, Dynamic Range, and Spectrum," *IEEE Trans. on Image Process.*, vol. 19, no. 9, pp. 2241–2253, Sept. 2010, doi: 10.1109/TIP.2010.2046811.

[37] B. Arad *et al.*, "NTIRE 2022 spectral recovery challenge and data set," in *2022 IEEE/CVF Conference on Computer Vision and Pattern Recognition Workshops (CVPRW)*, June 2022, pp. 862–880. doi: 10.1109/CVPRW56347.2022.00102.

[38] Z. Shi, C. Chen, Z. Xiong, D. Liu, and F. Wu, "HSCNN+: Advanced CNN-Based Hyperspectral Recovery from RGB Images," in *2018 IEEE/CVF Conference on Computer Vision and Pattern Recognition Workshops (CVPRW)*, June 2018, pp. 1052–10528. doi: 10.1109/CVPRW.2018.00139.

[39] J. Li, C. Wu, R. Song, Y. Li, and F. Liu, "Adaptive Weighted Attention Network with Camera Spectral Sensitivity Prior for Spectral Reconstruction from RGB Images," in *2020 IEEE/CVF Conference on Computer Vision and Pattern Recognition Workshops (CVPRW)*, June 2020, pp. 1894–1903. doi: 10.1109/CVPRW50498.2020.00239.

[40] Y. Cai *et al.*, "MST++: Multi-stage Spectral-wise Transformer for Efficient Spectral Reconstruction," in *2022 IEEE/CVF Conference on Computer Vision and Pattern Recognition Workshops (CVPRW)*, June 2022, pp. 744–754. doi: 10.1109/CVPRW56347.2022.00090.

[41] N. Ketkar, "Introduction to PyTorch," in *Deep Learning with Python: A Hands-on Introduction*, N. Ketkar, Ed., Berkeley, CA: Apress, 2017, pp. 195–208. doi: 10.1007/978-1-4842-2766-4_12.